\documentclass[runningheads]{llncs}

\usepackage{eccv}

\usepackage{eccvabbrv}

\usepackage{graphicx}
\usepackage{booktabs}
\usepackage{amsmath}
\usepackage{amssymb}
\usepackage{booktabs}
\usepackage{xfrac}
\usepackage{multirow}
\usepackage{amssymb}
\usepackage{pifont}
\newcommand{\cmark}{\ding{51}}%
\usepackage{colortbl}
\definecolor{Gray}{gray}{0.9}
\usepackage[accsupp]{axessibility}  

\usepackage{hyperref}

\usepackage{orcidlink}

\begin{document}
	
	
	\title{Unveiling Advanced Frequency Disentanglement Paradigm for Low-Light Image Enhancement} 
	
	\titlerunning{AFD-LLIE}
	
\author{Kun Zhou\inst{1,2*}\orcidlink{0000-0001-9592-6575} \and
Xinyu Lin\inst{1,2}\thanks{Equal contribution}\orcidlink{0000-0003-0455-6199} \and
Wenbo Li\inst{3}\orcidlink{0000-0003-4604-778X} \and 
Xiaogang Xu \inst{3}\orcidlink{0000-0002-7928-7336}  \\ 
Yuanhao Cai \inst{4}\orcidlink{0000-0002-8266-7102} \and 
Zhonghang Liu \inst{5}\orcidlink{0009-0007-6589-7196} \and
Xiaoguang Han \inst{1}\orcidlink{0000-0003-0162-3296} \and
Jiangbo Lu \inst{2}\orcidlink{0000-0002-0048-3140}}
	
	\authorrunning{K. Zhou et al.}
	
\institute{ $^1$CUHK-Shenzhen, China \quad \inst{2}SmartMore Corporation, China
  \quad \inst{3}CUHK \\  \inst{4}Johns Hopkins University, USA \quad \inst{5}SMU, Singapore\\
\email{jiangbo.lu@gmail.com} }
	
	\maketitle

	\begin{abstract}
		
		Previous low-light image enhancement (LLIE) approaches, while employing frequency decomposition techniques to address the intertwined challenges of low frequency (e.g., illumination recovery) and high frequency (e.g., noise reduction), primarily focused on the development of dedicated and complex networks to achieve improved performance. In contrast, we reveal that an advanced disentanglement paradigm is sufficient to consistently enhance state-of-the-art methods with minimal computational overhead. Leveraging the image Laplace decomposition scheme, we propose a novel low-frequency consistency method, facilitating improved frequency disentanglement optimization. Our method, seamlessly integrating with various models such as CNNs, Transformers, and flow-based and diffusion models, demonstrates remarkable adaptability. Noteworthy improvements are showcased across five popular benchmarks, with up to 7.68dB gains on PSNR achieved for six state-of-the-art models. Impressively, our approach maintains efficiency with only 88K extra parameters, setting a new standard in the challenging realm of low-light image enhancement. \url{https://github.com/redrock303/ADF-LLIE}.

		\keywords{Low-light Image Enhancement, Disentanglement Optimization, Frequency Consistency}
	\end{abstract}

	\section{Introduction}
	\label{sec:intro}
	
\begin{figure*}[t]
		\centering

		\includegraphics[width=1.0\columnwidth]{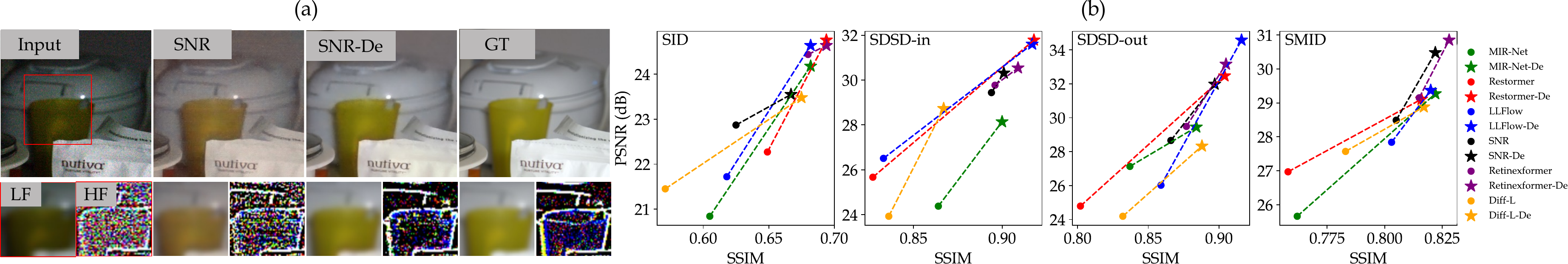} 
		\caption{(a) Illustration of predictions and corresponding frequency-disentangled components. `SNR-De' denote the improved version of SNR~\cite{xu2022snr} by our method.  Visualization of the decoupled components shows that our method accurately recovers intensity in the low-frequency domain and effectively denoises in the high-frequency domain (see supplementary materials for more visual results).  (b) Comparison between SOTA models and their improved versions on four representative benchmarks.  It can be seen that our method significantly improves six representative SOTA models. $\bullet$ and $\star$ denote these baseline (SOTA) models and their enhanced version by our method, respectively. }
		\label{fig:teasing_perform}

	\end{figure*}

	Low-light image enhancement (LLIE) endeavors to enhance the quality and exposure of photographs taken in dark environments. The state-of-the-art (SOTA) LLIE methods~\cite{xu2022snr,Cai_2023_ICCV,zamir2022restormer,guo2020zero,liang2023low,wang2022low,jiang2023low} achieve remarkable progress by leveraging advanced algorithms~\cite{wang2022low,xu2022snr,jiang2023low,hou23global} and various deep architectures~\cite{Cai_2023_ICCV,zamir2022restormer}. 
	These models generally address the combined challenges of low-frequency adjustment and high-frequency restoration within a unified framework. However, these two tasks exhibit distinct characteristics and differ from each other in several aspects. For example, low-frequency adjustments can unintentionally amplify noise~\cite{zheng2021adaptive,xu2020learning}, while high-frequency restoration may impact the recovery of illumination intensity. Relying on a single, unified model presents significant challenges due to the complexities~(combined low-frequency and high-frequency enhancement) involved in the optimization processes. Consequently, this routine may result in suboptimal outcomes, such as inaccurate illumination corrections or residual noise artifacts, as illustrated in Fig.~\ref{fig:teasing_perform}(a). 
	
	Given the challenges associated with coupled frequency optimization, a pertinent question arises: How can we devise a generic frequency disentanglement paradigm capable of (1) seamlessly integrating with existing LLIE methods, (2) enhancing their frequency restoration capabilities, and (3) requiring a minimal additional model complexity (e.g., extra parameters, computational costs)? In response to this query, we introduce a generic frequency disentanglement optimization paradigm aimed at augmenting previous LLIE models.
	
	Existing frequency decomposition learning schemes neglect the optimization interaction of these distinct frequency components and require addressing the combined degradations. In contrast, we take advantage of the image Laplace pyramid and decompose images into different frequency bands. More importantly, we present a novel low-frequency consistency constraint to decompose the complex combined restoration problem into two relatively simpler tasks, namely coarse low-frequency adjustment and frequency decoupled restoration. 
	
	In the coarse low-frequency adjustment, we present a novel Adaptive Convolutional Composition Aggregation module~(ACCA), focusing on challenges at low frequency. In ACCA, we design a convolution-style spatial-channel attention mechanism to perform effective content aggregation, largely reducing the overall complexity. It outperforms Retinexformer~\cite{Cai_2023_ICCV} with a substantial \textit{1dB} margin in PSNR on the LOL-v2~\cite{yang2020fidelity}, which is $5.5\%$ model size of Retinexformer (1.6M).

	In our frequency decoupled restoration, we offer the Laplace Decoupled Restoration Model (LDRM), which is mainly for high-frequency enhancement. By introducing a novel low-frequency consistency term, LDRM vastly reduces the complexity of optimization. As illustrated in Fig.~\ref{fig:teasing_perform} (b), our proposed approach reports remarkable improvement over SOTA image enhancement models, including MIR-Net~\cite{Zamir2020MIRNet}, Restormer~\cite{zamir2022restormer}, LLFlow~\cite{wang2022low}, SNR~\cite{xu2022snr}, Retinexformer~\cite{Cai_2023_ICCV} and Diff-L~\cite{jiang2023low}, all achieved with only a marginal increase in computational costs (\ie, 88K parameters and 2.53GFLOPS for $256\times256$ inputs in Tab.~\ref{tab:cost}). 
	Significantly, our method notably enhances Restormer by an impressive $6.11$dB, $7.68$dB in PSNR on the SDSD-in~\cite{wang2021seeing} and SDSD-out~\cite{wang2021seeing} benchmarks, respectively. In Fig.~\ref{fig:teasing_perform}(a) and Fig.~\ref{fig:sota}, it is evident that our approach significantly improves the visual quality of the restored predictions for SOTA models, effectively addressing illumination degradation and mitigating artifacts. 
	
	In summary, our contributions are threefold: \\
	
	\noindent {\textbf{Disentanglement learning.}}
	We introduce an effective decomposition learning framework, yielding a significant improvement. This strategy achieves separate optimization in the low-frequency and high-frequency domains while maintaining low-frequency consistency between the two stages.
	
	\noindent {\textbf{Efficient coarse adjustment.}}
	Our ACCA introduces adaptive spatial-channel aggregation, resulting in SOTA low-frequency adjustment results. Remarkably, it achieves competitive results while being parameter-efficient (Tab.~\ref{table:ccasota}).

	\noindent {\textbf{Universal improvements over SOTA models.}}
	Leveraging our disentanglement learning framework and lightweight coarse adjustment, we significantly improve prevailing LLIE models both quantitatively and qualitatively. Meanwhile, it allows effortless deployment of advanced low-frequency or high-frequency enhancement models as validated in Sec.~\ref{sec:exp}.

	\section{Related Works}
	\label{sec:relatedwork}
	\noindent\textbf{Traditional LLIE methods.} Traditional low-light image enhancement methods have been extensively explored to improve the visibility and quality of images captured under challenging lighting conditions. Histogram equalization~\cite{pizer1987adaptive} has been widely employed to enhance contrast, but it tends to amplify noise and lacks local adaptability. On the other hand, Retinex-based methods~\cite{jobson1997multiscale,feng2020low} have aimed to separate reflectance and illumination components, but they often struggle with low-light scenarios, leading to over-enhancement and unrealistic results. Contrast-limited Adaptive Histogram Equalization (CLAHE)~\cite{reza2004realization} attempts to address local contrast enhancement, but it may introduce halos around sharp edges, causing distortions and unnatural-looking enhancements.

	\noindent\textbf{Deep LLIE models.}
	Deep learning-based methods~\cite{wei2018deep,yang2021sparse,Chen2018Retinex,moran2020deeplpf,liu2021ruas,moran2021curl,hu2018exposure,park2018distort,gharbi2017deep,souibgui2022docentr,wang2022structural,kim2021representative,kim2020global} have gained significant attention for low-light image enhancement, aiming to tackle the challenges of both lighting improvement and denoising through a unified framework. Retinex-Net~\cite{wei2018deep} combines the benefits of traditional Retinex algorithms with convolutional neural networks (CNNs). It utilizes a multi-scale architecture to decompose the input low-light image into illumination and reflectance components.  MIRNet~\cite{Zamir2020MIRNet} utilizes a multi-scale architecture to capture and distill information at different levels. On the other hand, transformer-based architectures~\cite{xu2022snr,wang2022structural,zhang2021star,cui2022illumination,kim2021representative,zhou2022mutual,peng2021u,zamir2022restormer} have dominated the LLIE field with impressive performance. For example,  Star~\cite{zhang2021star} uses a lightweight transformer to learn coarse image structure information for image enhancement.  More recently, Retinexformer~\cite{Cai_2023_ICCV} presents a one-stage Retinex-based transformer for enhancing low-light images. These transformer-based models are generally good at long-range correlation modeling but require high computational costs. 
	
	\noindent\textbf{Decomposition-based LLIE.}
	There are two primary categories of decomposed LLIE methods~\cite{liang2023low,xu2020learning,hao2020low,fu2023you,liang2023low}. First, frequency-based image restoration has garnered significant attention, particularly within the LLIE field. Xu \etal~\cite{xu2020learning} propose a two-stage enhancement pipeline for low-light image enhancement, where they introduce a unified framework to optimize the main backbone module across both stages. Secondly, Retinex-based approaches represent low-light and normal-light images using two key components: reflectance and illumination layers. These methods estimate the reflectance as the predicted enhancement result. Hao \etal~\cite{hao2020low} present a semi-decoupled learning method to restore both layers using different optimization terms. RFR~\cite{fu2023you} introduces a self-knowledge distillation method for Retinex decomposition based on contrastive learning.
	
	In summary, lots of frequency-based decomposition models have explored various techniques for LLIE. Though they explicitly decompose images into different frequency bands, they \textit{do not decouple the optimization process} of low-frequency and high-frequency components.
	In contrast, our decoupled frequency restoration framework, featuring a new low-frequency consistent loss, establishes a universal decoupled enhancement paradigm that enhances various baselines.

	\section{Method}
	\label{sec:method}
	As mentioned previously, LLIE aims to address issues of coupled low-frequency (e.g., illumination recovery) and high-frequency (e.g., noise reduction) enhancement.
	Different from existing LLIE models utilizing a unified solution to handle diverse (combined) conditions, our research delves into the potential advantages of a decomposition learning strategy for image enhancement.
	
	\begin{figure*}[ht]
		\centering
		\includegraphics[width=1\columnwidth]{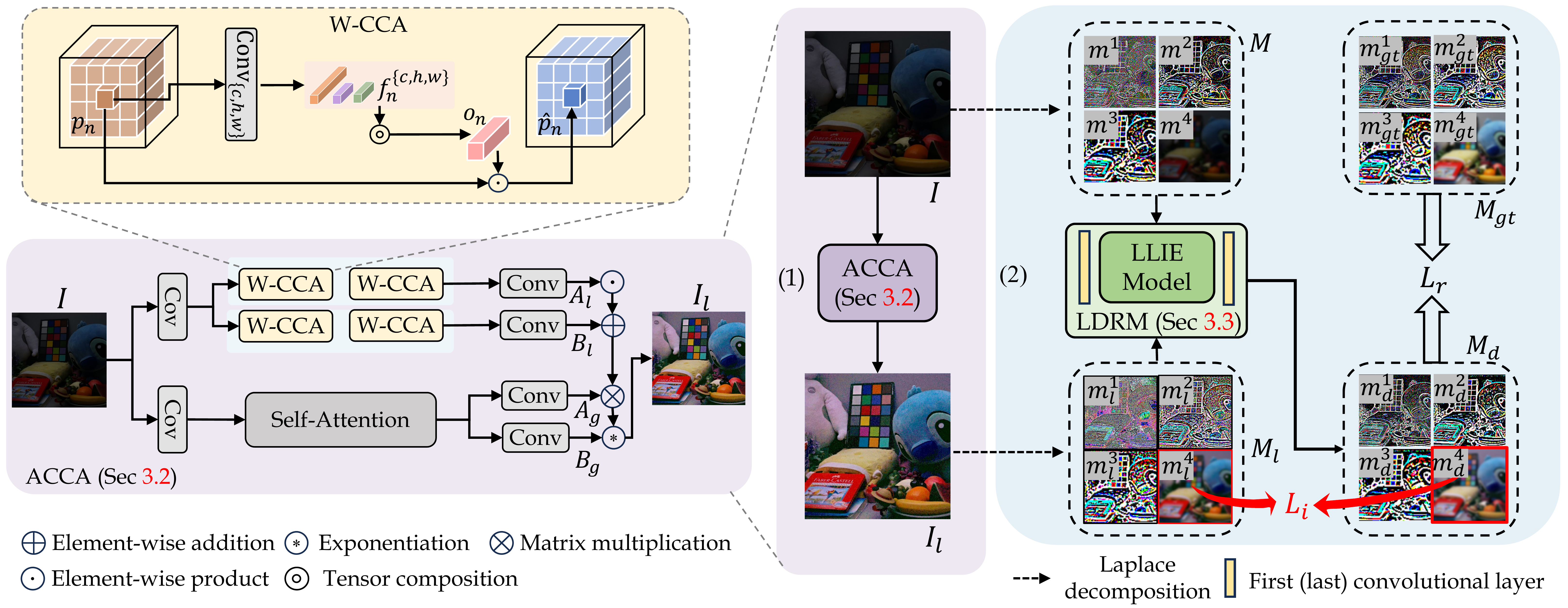} 
		\caption{Overview of our proposed frequency disentanglement learning framework. It consists of two phases: (1) coarse phase: ACCA conducts coarse adjustment to initially enhance the input image $I$ and produce the preliminary result $I_l$, and (2) coarse-to-fine phase: LDRM integrates Laplace representations~($M, M_l$ from the $I, I_l$) for subsequent fine-grained restoration. Additionally, a low-frequency consistent loss ${\color{red}L_i}$~(Eq.~\ref{eq:loss}) between the two phases is introduced to achieve effective disentanglement optimization.}
		\label{fig:framework}
	\end{figure*} %
	
	Based on the observation that enhancement in the low-frequency domain reports a greater impact and thus encourages lightweight structures to favor coarse learning, we design a lightweight coarse adjustment mainly for low-frequency degradation (Sec.~\ref{sec:cfa}) and a consistent coarse-to-fine~(C2F) restoration (Sec.~\ref{sec:laplace}) principally for high-frequency issues, as illustrated in Fig.~\ref{fig:framework}.
	Initially, we present the Adaptive Convolutional Composition Aggregation module (ACCA) to provide coarse adjustments, primarily focusing on illumination recovery. Following this, we utilize the Laplace Decoupled Restoration Model (LDRM) to separate the coarse results and original inputs into multi-scale high and low-frequency maps. It's worth noting that LDRM seamlessly integrates with state-of-the-art LLIE models with minimal redesign. Lastly, our proposed low-frequency consistent supervision ensures efficient decomposition learning.

	\subsection{Overview}
	Our goal is to recover a high-quality output image $I_c$ from a low-light noisy image $I$. 
	In general, we disentangle the prevailing coupled optimization into coarse phase and coarse-to-fine phase. During the coarse adjustment, we employ the ACCA module to mainly recover the low-frequency information in the image $I$, obtaining the roughly enhanced results $I_l$ that usually have better brightness:
	\begin{equation}
		I_l =  f_l(I,\theta_l),
	\end{equation}
	where $\theta_l$ is the learnable parameters of the ACCA module.
	
	Subsequently, we proceed to the coarse-to-fine phase. In this stage, with our primary emphasis on the high-frequency domain, we propose a Laplace decoupled representation to separate low-frequency and high-frequency maps from both the original input $I$ and the coarsely recovered image $I_l$:
	\begin{equation}
		M =  {\rm Lap}(I), M_l = {\rm Lap}(I_l),
		\label{eq:lap}
	\end{equation}
	where $M=[m^1,m^2,\cdots,m^K],M_l=[m^1_l,m^2_l,\cdots,m^K_l]$ refers to the decomposed maps of $I$, $I_l$, and $K$ is the pyramid level. The coarsest feature map $m^K, m^K_l$ generally contains the most low-frequency information of the images $I$ and $I_l$, while the other feature maps $\{m^1,\cdots m^{K-1}, m^1_l,\cdots m^{K-1}_l\}$ capture the higher-frequency nuances. Following an upsampling operation on all the $K-1$ low-resolution maps, we stack all of these maps alone with the channel dimension:
	
	\begin{equation}
		M_s =  {\rm Stack}[m^1,{\rm Up}(m^2),\cdots,{\rm Up}(m^K),
		m^1_l,{\rm Up}(m^2_l),\cdots,{\rm Up}(m^K_l)],	
	\end{equation}
	where ``${\rm Up}$'' means bilinear up-sampling to up-scale all the low-resolution maps to the same size with the first map $m^1$. Finally, Our proposed LDRM takes the stacked maps $M_s$ as input and restores the enhanced Laplace maps $M_d=[m^1_d,m^2_d,\cdots,m^K_d]$:
	\begin{equation}
		M_d =  f_d(M_s,\theta_d),
	\end{equation}
	where $\theta_d$ is the learnable parameter of LDRM.
	To build effective supervision, we first use the reconstruction loss $L_{r}$ to guide the multi-scale predictions $M_d$ closer to corresponding decomposed maps of ground truth $M_{gt}$ obtained by the same Laplace decouple pipeline.
	In addition, we also introduce a low-frequency consistent supervision $L_{i}$ to effectively decompose the coarse low-frequency adjustment and high-frequency refine learning. The overall loss of LDRM is:
	
	\begin{equation}
		L_{total} = L_{r} + \alpha * L_{i}, 
		L_{r}  = \sum_{i=1}^{K}(\displaystyle || m^i_d-m_{gt}^i ||_1^1),
		L_{i}= \displaystyle || m^K_d-m_{l}^K ||_1^1,
		\label{eq:loss}
	\end{equation}
	where $\alpha$ is the scale factor to balance these two loss terms, and $\displaystyle || \cdot ||_1^1$ refers to $\ell_1\text{-norm}$. 
	In the next section, we will explain ACCA for the coarse learning phase and Laplace decomposition schemes for coarse-to-fine restoration, respectively.
	
	\subsection{ACCA for Coarse Adjustment}
	\label{sec:cfa}
	
	{\noindent\bf{Framework overview.}}
	Inspired by~\cite{cui2022illumination}, we utilize a hybrid dual-branch neural network aimed at regressing both global and local low-frequency adjustment parameters. As shown in Fig.~\ref{fig:framework}, our proposed Adaptive Convolutional Composition Aggregation (ACCA) is composed of two branches. 
	In the local branch, we first employ a convolution layer to get a corresponding deep image feature of the input image. Subsequently, two parallel Window-based Convolutional Composition Attention blocks (W-CCA) are developed to regress two local transformation factor maps ${\{A_l,B_l\}}$, where $A_l\in R^{\{H\times W\times 3\}}$ is a scale amplification map and $B_l\in R^{\{H\times W\times 3\}}$ is a linear compensation map. The result of local parameters adjustment is calculated as:
	\begin{equation}
		I_l^{local} = A_l \odot I \oplus B_l, 
	\end{equation}
	where $\odot$ refers to element-wise multiplication and $\oplus$ means the element-wise addition.
	Simultaneously, we also adopt a global ISP (Image Signal Processing) branch\footnote{The structure of our global branch is illustrated in our supplementary material.} that employs a transformer-based framework, achieving a global enhancement for the locally adjusted result:
	\begin{equation}
		I_l = (A_g \otimes I_l^{local})^{B_g},
	\end{equation}
	where $A_g,B_g$ are the predicted colour matrix and gamma adjustment parameters and $\otimes$ is the matrix multiplication.

	{\noindent\bf{W-CCA.}}
	Exiting studies~\cite{wang2020eca,wang2023omni} have proved the importance of both spatial and channel aggregation for some low-level tasks, including denoising, super-resolution, deblurring, colorization, etc. However, performing spatial-channel fusion encounters challenges due to the high computational demands. To tackle this issue, we introduce a novel window-based convolutional composition method for efficient local enhancement.
	
	Given a 2D image feature $F$, we first employ a stridden convolutional layer to split $F$ into several non-overlapped image patches: $P = {\rm Conv}(F,s,s)$.
	
	$\rm Conv(\cdot)$ denotes a grouped convolution with group size and stride size $s$. As a result, $F$ is partitioned into $N=H\times W/s^2$ patches $P=[p_1,p_2,\cdots,p_N]$.
	For $n$-th divided image patch $p_n\in R^{\{s\times s\times C\}}$, where $C$ represents the channel dimension and $s$ denotes the window size. Then we perform Omni aggregation using a 3D similarity map $o_n\in R^{\{s\times s\times C\}}$:
	\begin{equation}
		{\hat p_n} = o_n \odot p_n.
		\label{eq:oagg}
	\end{equation}
	Instead of directly regressing the corresponding 3D Omni similarity map $o_n$, we employ the tensor composition technique~\cite{kolda2009tensor,kuleshov2015tensor,wu2018neural} to produce $o_n$ by three regressed 1D tensors:
	\begin{equation}
		o_n = f^h_n\circledcirc \ f^w_n \circledcirc f^c_n.
		\label{eq:fcl}
	\end{equation}
	The terms $f^h_n$ and $f^w_n$ represent two separable tensors in 2D space, specifically referring to the height and width dimensions. The operation $\circledcirc$ denotes the composition of these tensors. Additionally, the term $f^c_n$ represents a 1D tensor that is used to establish correlations along the channel dimension. 
	Towards a lower computational cost and pixel-adaptive aggregation, as depicted in Fig.~\ref{fig:framework}, triple convolutions with a stride of $s$ are utilized to regress the three separable kernels:
	\begin{equation}
		f^{\{h,w,c\}}_n = {\rm Conv_{\{h,w,c\}}}(p_n,s,s).
		\label{eq.deconv}
	\end{equation}
	To summarize, we derive 3D Omni similarity maps by Eq.~\ref{eq:fcl} and perform Omni similarity aggregation by Eq.~\ref{eq:oagg}.

	{\noindent\bf{Complexity analysis of W-CCA.}}
	In this part, we will show the significant reduction in computational resource consumption achieved by W-CCA.
	As can be seen in the last paragraphs, we employ a stridden convolution layer to split an input image feature $F$ into $M_h\times M_w$ image patches\footnote{$M_h = H/s, M_w = W/s$, $H\times W$ is the feature resolution.}. To further reduce the parameter count, we implement a group-wise convolution with a group size of $s$ and a kernel size of $s \times s$. The total computational complexity of a single W-CCA is determined by the cumulative sum of operations in Eq.~\ref{eq:oagg}, Eq.~\ref{eq:fcl}, Eq.~\ref{eq.deconv} and the first stridden convolution:
	\begin{equation}
		\begin{split}
			\mathcal{O}(\text{W-CCA}) = 4HWC + 2HWC^2/s\;.
		\end{split}	
		\label{eq.complexity}
	\end{equation}
	A more detailed complexity analysis is reported in supplementary material.
	
	From the Eq.~\ref{eq.complexity}, we can see the complexity of our proposed W-CCA is {\textit{linearly}} increased with the image resolution $H$ and $W$.

	\subsection{LDRM for Consistent C2F Restoration}
	\label{sec:laplace}
	In the Laplace Decoupled Restoration Model (LDRM), we re-design SOTA existing LLIE models by integrating Laplace representation (also known as Laplace pyramid~\cite{burt1987laplacian,adelson1984pyramid,li2021multi,liang2021high}) for subsequent effective fine-grained enhancement. In general, we only modify the first and last convolutional layers to employ our Laplace pyramid representation, dissecting a given image into multi-scale high-frequency components and a low-frequency part.

	To begin with, we first utilize a Laplace pyramid decomposition layer to obtain multi-scale image maps from both input image $I$ and the coarse enhanced result $I_l$. Taking the input image $I$ for illustration, we follow~\cite{burt1987laplacian} and obtain the Gaussian pyramid:
	\begin{equation}
		{I}_G^k =
		\begin{cases}
			g \ast I \;; & \operatorname{if} k=1\\
			g \ast {\rm resize}({I}_G^k,\downarrow2) ; & \operatorname{if}  \ k>1\\
		\end{cases}
	\end{equation}
	where $g(\cdot)$ refers to a $s\times s$ Gaussian kernel, $\ast$ means the convolutional operation and `${\rm resize}$' is a bilinear interpolation. Then, We derive a Laplacian pyramid by computing the difference between two neighboring Gaussian maps:
	\begin{equation}
		m^k =
		\begin{cases}
			I - {I}_G^k\;; & \operatorname{if} k=1\\
			{\rm resize}({I}_G^{k-1},\downarrow2) - {I}_G^k\;; & \operatorname{if} K>k>1\\
			{\rm resize}({I}_G^{k-1},\downarrow2) ; & \operatorname{if}  \ k=K\\
		\end{cases}
		\label{eq:lapde}
	\end{equation}
	As shown in Eq.~\ref{eq:lapde}, the high-frequency components are produced by the DoG~(Difference of Gaussians).
	
	After that, all these decomposed Laplace maps are stacked and fed into our LDRM to generate the corresponding restored Laplace maps. The final reconstructed output is obtained by inverse Laplace transformation:
	\begin{equation}
		{\hat m}^{k} = 
		\begin{cases}
			m^{K}; & \operatorname{if} k=K \\
			m^k + {\rm resize}(m^{k+1},\uparrow2); &\text{otherwise} \\
		\end{cases}	
		\label{eq.iterativelap}
	\end{equation}
	where ${\hat m}^1$ is the final reconstructed image, and the inverse Laplace decouple is performed in reverse order, from ${\hat m}^{K}$ to ${\hat m}^{1}$. Note that we set $K=4$ in our work, for it outperforms other $K$ settings.

	\section{Experiments}
	\label{sec:exp}
	\subsection{Settings}
	
	{\noindent\bf{Datasets.}}
	We evaluate our approach on several datasets for low-light image enhancement, including LOL-v2~\cite{yang2020fidelity}, SID~\cite{chen2019seeing}, SDSD~\cite{wang2021seeing} (both indoor and outdoor parts), and SMID~\cite{chen2018learning}. 
	LOL-v2 dataset: we specifically focus on its real subset, which presents challenging low-light conditions and noise degradations. This subset comprises 689 training samples and 100 testing samples.  
	SID dataset: we utilize 2099 images for training, while the remaining 598 image pairs are reserved for evaluation.    
	SDSD-in/SDSD-out datasets: these datasets consist of static indoor and outdoor scenes captured in various environments.    
	SMID dataset: it contains static scenes captured in outdoor environments. We follow previous SOTA methods and train/evaluate various baseline models and their counterparts (with our disentanglement optimization) on the training part of each dataset, \textit{ensuring the fairness comparison and evaluation.}

	\begin{table*}[t]
		\centering
		\scriptsize
		\setlength{\tabcolsep}{3pt}
		
		\renewcommand{\arraystretch}{1.0}
		\caption{
			Quantitative results (PSNR (dB)/SSIM) on five challenging LLIE benchmarks. `-De' refers to the enhanced version of our proposed approach and `Improve.' refers to the improvements attained by our method compared to the six baselines. 
		}
		\label{tab:psnr-ssim}
		\begin{tabular}{l|ccccc}
			\hline 
			\multirow{2}{*}{Method}& LOL-v2 & SID & SDSD-in & SDSD-out  &SMID  \\
			\cline{2-6}
			& PSNR/SSIM & PSNR/SSIM & PSNR/SSIM & PSNR/SSIM & PSNR/SSIM  \\
			\hline
			MIR-Net~\cite{Zamir2020MIRNet}    & 20.02/0.820 & 20.84/0.605 & 24.38/0.864 & 27.13/0.837 &25.66/0.762   \\
			
			\rowcolor{gray!10} MIR-Net-De    & 24.19/0.882 &24.18/0.682  & 28.14/0.900 & 29.45/0.884 & 29.27/0.822 \\ 
			\rowcolor{gray!10} Improve.	 	& {\color{red}{+4.17}/{+0.062}} &{\color{red}{+3.34}/{+0.075}}  & {\color{red}{+3.76}/{+0.036}} & {\color{red}{+2.32}/{+0.047}}  & {\color{red}{+1.16}/{+0.072}} \\ \hline 
			
			Restormer~\cite{zamir2022restormer}        & 19.94/0.827 & 22.27/0.649 & 25.67/0.827 & 24.79/0.802 &26.97/0.758   \\
			
			\rowcolor{gray!10}  Restormer-De       & 24.56/0.893 &24.76/0.694  & 31.78/0.918 & 32.47/0.904 & 29.11 /0.816  \\ 
			\rowcolor{gray!10}	Improve.		& {\color{red}{+4.62}/{+0.066}} &{\color{red}{+2.49}/{+0.045}}  & {\color{red}{+6.11}/{+0.091}} & {\color{red}{+7.68}/{+0.102}} & {\color{red}{+2.14}/{+0.058}}  \\ \hline 
			LLFlow~\cite{wang2022low}     &26.20/0.888 & 21.72/0.618 & 26.51/0.883 & 26.02/0.859 & 27.84/0.803   \\
			
			\rowcolor{gray!10} LLFlow-De    & 28.90/0.908 & 24.64/0.682  & 31.60/0.917 & 34.58/0.916 & 29.37/0.820  \\ 
			\rowcolor{gray!10} Improve.		& {\color{red}{+1.70}/{+0.020}} &{\color{red}{+2.92}/{+0.064}}  & {\color{red}{+5.09}/{+0.034}} & {\color{red}{+8.83}/{+0.057}} & {\color{red}{+1.53}/{+0.017}}  \\ 
			\hline 
			
			SNR~\cite{xu2022snr}    & 21.48/0.849 & 22.87/0.625 & 29.44/0.894 & 28.66/0.866 & 28.49/0.805   \\
			
			\rowcolor{gray!10} SNR-De      & 24.00/0.872 & 23.55/0.667  & 30.31/0.901 & 31.98/0.897 & 30.48/0.822 \\ 
			\rowcolor{gray!10}	Improve.		& {\color{red}{+2.52}/{+0.023}} &{\color{red}{+0.68}/{+0.042}}  & {\color{red}{+0.87}/{+0.007}} & {\color{red}{+3.32}/{+0.031}} & {\color{red}{+1.99}/{+0.017}} \\ \hline 
			
			Retinexformer~\cite{Cai_2023_ICCV}     & 22.80/0.840 & 24.44/0.680 & 29.77/0.896 & 29.49/0.877 & 29.15/0.815   \\
			
			\rowcolor{gray!10} Retinexformer-De    & 24.21/0.881 & 24.64/0.694  & 30.54/0.909 & 33.16/0.905 & 30.85 /0.828  \\ 
			\rowcolor{gray!10} Improve.		& {\color{red}{+1.41}/{+0.041}} &{\color{red}{+0.20}/{+0.014}}  & {\color{red}{+0.77}/{+0.013}} & {\color{red}{+3.67}/{+0.028}} & {\color{red}{+1.70}/{+0.013}}  \\ \hline 
			
			Diff-L~\cite{jiang2023low}     & 18.95/0.722 & 21.45/0.571 & 23.93/0.836 & 24.19/0.832 & 27.57/0.783    \\
			
			\rowcolor{gray!10} Diff-L-De    & 23.93/0.853 & 23.48/0.675  & 28.73/0.867 & 28.33/0.888 & 28.88/0.817  \\ 
			\rowcolor{gray!10} Improve.		& {\color{red}{+4.98}/{+0.081}} &{\color{red}{+2.03}/{+0.104}}  & {\color{red}{+4.80}/{+0.031}} & {\color{red}{+4.14}/{+0.056}} & {\color{red}{+1.31}/{+0.034}}  \\ 
			
			\hline
		\end{tabular}
	\end{table*}
	
	\begin{table}[h]
		\footnotesize
		\centering
		\setlength{\tabcolsep}{2pt}
		\caption{
			We present comprehensive details regarding model complexities, including parameters, GFLOPS, and inference time. Consistent with~\cite{Cai_2023_ICCV}, we assess GFLOPS and inference time using an input size of $256\times256$, utilizing an NVIDIA-RTX 3090 GPU. It's worth highlighting that our framework~(denoted as `w/ ours') entails only $0.2\%$-$5.5\%$ additional parameters for enhancing the state-of-the-art LLIE models.
		}
		\label{tab:cost}
		\begin{tabular}{l|c|c|c|c|c|c||c } 
			\hline
			Methods &MIR-Net &Restormer &LLFlow &SNR &Retinexformer  &Diff-L    &w/ ours\\ \hline
			Param.(M) &31.79 &26.13 &37.68 &39.12 &1.61 &22.08 &+0.088\\
			GFLOPS &785 &144.25 &287 &26.35  &15.57 &88.92  &+2.53\\
			Speed(s)  &0.205 &0.104  &0.267 &0.039  &0.079 &0.227 &+0.008\\ \hline
		\end{tabular}
		
	\end{table}
	
	{\noindent\bf{Evaluation metrics.}}
	Following the evaluation of most LLIE approaches~\cite{Zamir2020MIRNet,zamir2022restormer,xu2022snr,Cai_2023_ICCV}, we assess the effectiveness of our method using the PSNR(${\color{red}\uparrow}$) (Peak Signal-to-Noise Ratio) and SSIM~\cite{wang2004image}(${\color{red}\uparrow}$) (Structural Similarity Index Measure) metrics. Higher values indicate better results.

	{\noindent\bf{Implementation details.}} 
	Our framework is implemented in PyTorch~\cite{pytorch}. We use common data augmentation techniques, including random cropping, vertical/horizontal flipping, and rotation. The Adam~\cite{kingma2014adam} optimizer and cosine annealing scheme~\cite{loshchilov2016sgdr} dynamically adjust the learning rate from $5\times 10^{-4}$ to 0.
	We initially train the ACCA module for 100 epochs on a single NVIDIA RTX 3090 GPU. Then, we freeze the ACCA weights and train the LDRM for 40K iterations on two NVIDIA RTX 3090 GPUs. The batch sizes are set to 1 (ACCA) and 32 (LDRM), respectively. All original baseline models and their corresponding LDRMs are \textbf{\textit{trained from scratch}}.
	To achieve decoupled high-frequency restoration, we propose a novel low-frequency consistent loss in our coarse-to-fine enhancement, as illustrated in Eq.~\ref{eq:loss}. Following IAT\cite{cui2022illumination}, the ACCA module is trained using L1 loss and VGG perceptual loss~\cite{johnson2016perceptual}.
	
	
	\subsection{Improvement over SOTA LLIE Models}
	We conduct extensive experiments to assess our frequency-based disentanglement optimization. We integrate our learning strategy on six SOTA LLIE models: MIR-Net~\cite{Zamir2020MIRNet}, Restormer~\cite{zamir2022restormer}, LLFlow~\cite{wang2022low}, SNR~\cite{xu2022snr}, Retinexformer~\cite{Cai_2023_ICCV} and Diff-L~\cite{jiang2023low}, which serve as our baseline models. 
	The modified versions of these models, equipped with our strategy, are denoted as MIR-Net-De, Restormer-De, LLFlow-De, SNR-De, Retinexformer-De, and Diff-L-De, respectively.
	
	{\noindent\bf{Quantitative comparison.}}
	Tab.~\ref{tab:psnr-ssim} presents a quantitative comparison between the six baseline models and their enhanced counterparts using our disentanglement learning approach. It clearly demonstrates that our algorithm significantly improves the performance of all baseline models across all six benchmarks. Notably, our method achieves a remarkable enhancement of up to $7.68$dB ($31$\% improvement) for Restormer~\cite{zamir2022restormer} on the SDSD-out benchmark. When evaluated on the same dataset, our approach showcases a substantial improvement, surpassing Restormer~\cite{zamir2022restormer} by over $0.1$ in terms of SSIM. This improvement highlights the exceptional restoration of our proposed LDRM in the high-frequency domain.  Moreover, our decomposition representation leads to universal improvements for all popular frameworks, including CNNs, Transformers, flow-based, and diffusion-based methods. Even MIR-Net~\cite{zamir2022restormer}, an early SOTA LLIE model, achieves top-ranked results when enhanced with our algorithm.
	
	Furthermore, we report the model complexity of these six LLIE models and the extra cost introduced by our approach. The results in Tab.~\ref{tab:cost} indicate that the re-designed models require only an additional $0.2$\% to $5$\% in terms of model parameters and FLOPS, demonstrating the efficiency of our method.

	\begin{table}[h]
		\footnotesize
		\centering
		\setlength{\tabcolsep}{2pt}
		\caption{
			Quantitative comparison (PSNR (dB)/SSIM) between our proposed frequency disentanglement scheme and segmentation-guided LLIE strategy~\cite{wu2023skf}. * means segmentation information is required.  Note that `{\color{red}+}' (`{\color{green}-}') highlighted in {\color{red}red} ({\color{green}green}) denotes performance {\color{red}improvement} ({\color{green}reduction}).
		}
		\begin{tabular}{l|c|c|c||c|c|c } 
			\hline
			LOL-v2 &SNR &SNR-SKF$^*$ &SNR-De &LLFlow &LLFlow-SKF$^*$  &LLFlow-De    \\ \hline
			PSNR $\uparrow$ &21.48 &21.93(\textcolor{red}{+0.45}) &24.00(\textcolor{red}{+2.52}) & 26.20 & 28.45(\textcolor{red}{+2.25})  &28.90(\textcolor{red}{+2.70})  \\ \hline   
			SSIM $\uparrow$ &0.849 &0.845(\textcolor{green}{-0.006}) &0.872(\textcolor{red}{+0.023}) & 0.888 & 0.905(\textcolor{red}{+0.017})  &0.908(\textcolor{red}{+0.020})  \\ \hline   
			Param.(M) &39.12 &39.44(\textcolor{black}{+0.32}) &39.21(\textcolor{black}{+0.088}) &37.68  &39.91(\textcolor{black}{+2.23}) &37.77(\textcolor{black}{+0.088})   \\ \hline
		\end{tabular}
		\label{tab:skf}
	\end{table}
	
	{\noindent\bf{Comparison with segmentation guided LLIE.}}
	In previous arts, Wu \etal~\cite{wu2023skf} explored utilizing semantic prior for guidance LLIE~(named as ``SKF"), leading to improved performance with SOTA LLIE methods. Different from their works, we present a decoupled optimization scheme to enhance existing approaches. The detailed comparisons between SKF and ours are presented in Table~\ref{tab:skf}. It is clear that our proposed method attains superior improvements than SKF, with fewer extra model complexities. It is noted that our method does not require segmentation information.
	
	\begin{table}[t]
		\small 
		\caption{Quantitative comparison between ACCA and other SOTA LLIE models on LOL-v2~\cite{yang2020fidelity} benchmark. The best and second-best results are highlighted in \textcolor{red}{\textbf{red}} and \textcolor{blue}{\textbf{blue}}, respectively. We test the inference speed (FPS: Frames Per Second) with an input size of $256\times 256$ on an NVIDIA RTX 3090 GPU. }
		\label{table:ccasota}
		\begin{center}
			\begin{tabular}{l|c|cccc } 
				\hline
				Method & Venue	&PSNR $\uparrow$ &SSIM 	$\uparrow$& Params. $\downarrow$	&FPS  $\uparrow$  \\ 
				\hline
				ZeroDCE~\cite{guo2020zero} &  CVPR 2020 	&17.63&0.617         &\textbf{\color{blue}0.079M} 		  &62.50                          \\ 
				Star~\cite{zhang2021star} & ICCV 2021	&18.26&0.546          &\textbf{\color{red}0.027M} 		  & 58.82                     \\ 
				IAT~\cite{cui2022illumination}  & BMVC 2022 	&{\color{blue}\textbf{23.50}}&0.824         &0.091M 		  &  45.45                    \\ 
				PairLIE~\cite{fu2023learning} & CVPR 2023 	&18.80&0.721          &0.342M 		  &                55.56          \\ 
				\rowcolor{gray!10}	ACCA (Ours) & ECCV 2024   &{\color{red}\textbf{23.80}}&0.829         &0.088M 		  &{\color{red}\textbf{125}}                          \\ \hline
				
				MIR-Net~\cite{Zamir2020MIRNet} & ECCV 2020  &20.02&0.820          &31.79M 		  &               4.87         \\
				MIR-Net-v2~\cite{Zamir2022MIRNetv2} & T-PAMI 2022  &21.10 &0.821          &5.86M 		  & 15.6         \\
				Restormer~\cite{zamir2022restormer} & CVPR 2022  &19.94&0.827         &26.13M 		  &          9.62               \\
				SNR~\cite{xu2022snr}  & CVPR 2022  &21.48&{\color{red}\textbf{0.849}}         &39.12M 		  &  {\color{blue}\textbf{83.33}}                      \\
				Retinexformer~\cite{Cai_2023_ICCV} & ICCV 2023  &22.80&{\color{blue}\textbf{0.840}}          &1.61M 		  &   27.03                     \\ 
				LLformer~\cite{wang2023ultra} &AAAI 2023  &20.69 &0.759          &24.55M 		  & 8.85 \\
				Diff-L~\cite{jiang2023low}  & Sig. Asia 2023  &18.96&0.723          &22.08M 		  &    4.41                     \\
				\hline
			\end{tabular}
		\end{center}
	\end{table}

	\begin{table}[t]
		\footnotesize 
		\caption{Improvements over Restormer~\cite{zamir2022restormer} when integrating existing fast LLIE models and our proposed ACCA into our disentanglement learning framework. }
		\label{subtab:llieinter}
		\begin{center}
			\begin{tabular}{l|c||c|c|c|c|c } 
				\hline
				\multirow{2}{*}{Method}                      & Restormer  &w/  ZeroDCE &w/ Star &w/ PairLIE &w/ IAT &w/ ACCA    \\ 
				&(Baseline) &\& $L_i$ &\& $L_i$ &\& $L_i$ &\& $L_i$ &\& $L_i$ \\ \hline
				
				PSNR (dB) &19.94 &21.53 &22.40 &22.36 &23.97 &24.56\\
				\rowcolor{gray!10}
				PSNR Gain &- &\textcolor{red}{+1.59} &\textcolor{red}{+2.46} &\textcolor{red}{+2.42} &\textcolor{red}{+4.03} &\textcolor{red}{+4.62}\\ \hline
				SSIM &0.827 &0.861 &0.866 &0.860 &0.887 &0.893 \\ 
				\rowcolor{gray!10}
				SSIM Gain &- &\textcolor{red}{+0.034} &\textcolor{red}{+0.039} &\textcolor{red}{+0.033} &\textcolor{red}{+0.060} &\textcolor{red}{+0.066} \\ \hline	
			\end{tabular}
		\end{center}
	\end{table}

	{\noindent\bf{Effectiveness of ACCA.}}
	In this study, we propose an efficient Adaptive Convolutional Composition Aggregation (ACCA) for coarse adjustment. To assess the effectiveness of ACCA, we quantitatively compare it with state-of-the-art LLIE models~(lightweight models: ZeroDCE~\cite{guo2020zero}, Star~\cite{song2021starenhancer}, IAT~\cite{cui2022illumination}, PairLIE~\cite{fu2023learning}; large models:  MIR-Net~\cite{Zamir2020MIRNet}, MIR-Net-v2~\cite{Zamir2022MIRNetv2}, Restormer~\cite{zamir2022restormer}, SNR~\cite{xu2022snr}, Retinexformer~\cite{Cai_2023_ICCV}), LLformer~\cite{wang2023ultra} and Diff-L~\cite{jiang2023low}  on the LOL-V2. From the results in Tab.~\ref{table:ccasota}, it is evident that ACCA surpasses both lightweight and parameter-intensity LLIE methods in terms of PSNR and achieves competitive SSIM value compared with larger LLIE models, highlighting its superior performance.
	
	{\noindent\bf{Integration of other coarse models.}}
	We integrate lightweight LLIE methods—Star, ZeroDCE, PairLIE, and IAT—into our disentanglement learning framework to replace ACCA. The results in Tab.~\ref{subtab:llieinter} show that these models significantly enhance the SOTA LLIE method, Restormer. Notably, IAT\cite{cui2022illumination} improves Restormer by 4.03dB in PSNR and 0.06 in SSIM, demonstrating the effectiveness of our disentanglement strategy. This experiment underscores our framework's flexibility, allowing seamless integration of existing coarse adjustment algorithms.
	
\begin{figure*}[t]
		\centering
		
		\includegraphics[width=1\columnwidth]{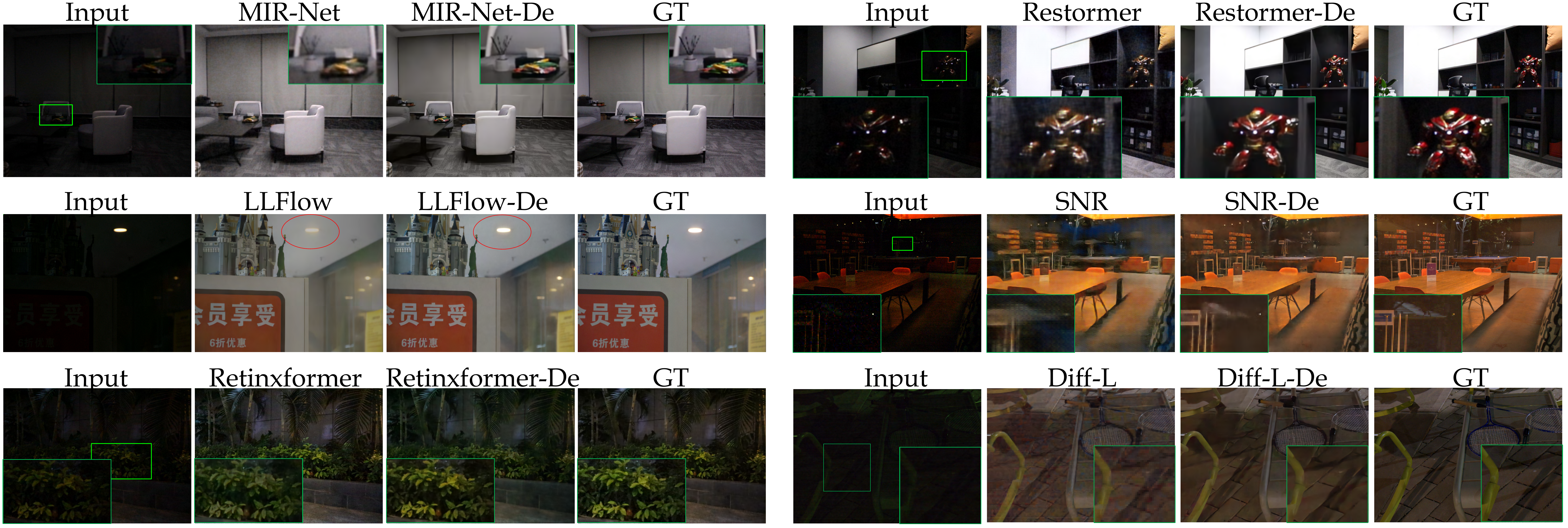} 
		\caption{Qualitative evaluation on several benchmarks. Our integration provides accurate outcomes for both high-frequency~(clearer image detail restoration) and low-frequency~(more accurate illumination recovery) areas.}

		\label{fig:sota}
		
	\end{figure*}

	{\noindent\bf{Qualitative comparison.}}
	The qualitative comparisons are depicted in Fig.~\ref{fig:sota}. It is evident that the SOTA models, primarily providing uniformly coupled optimization solutions, struggle to restore illumination and eliminate noise effectively. In contrast, with the integration of our strategy, they are capable of yielding more accurate outcomes with improved image details and fewer artifacts. More visual results are shown in our supplementary materials.

	\begin{figure}[t]
		
		\centering
		
		\includegraphics[width=1\columnwidth]{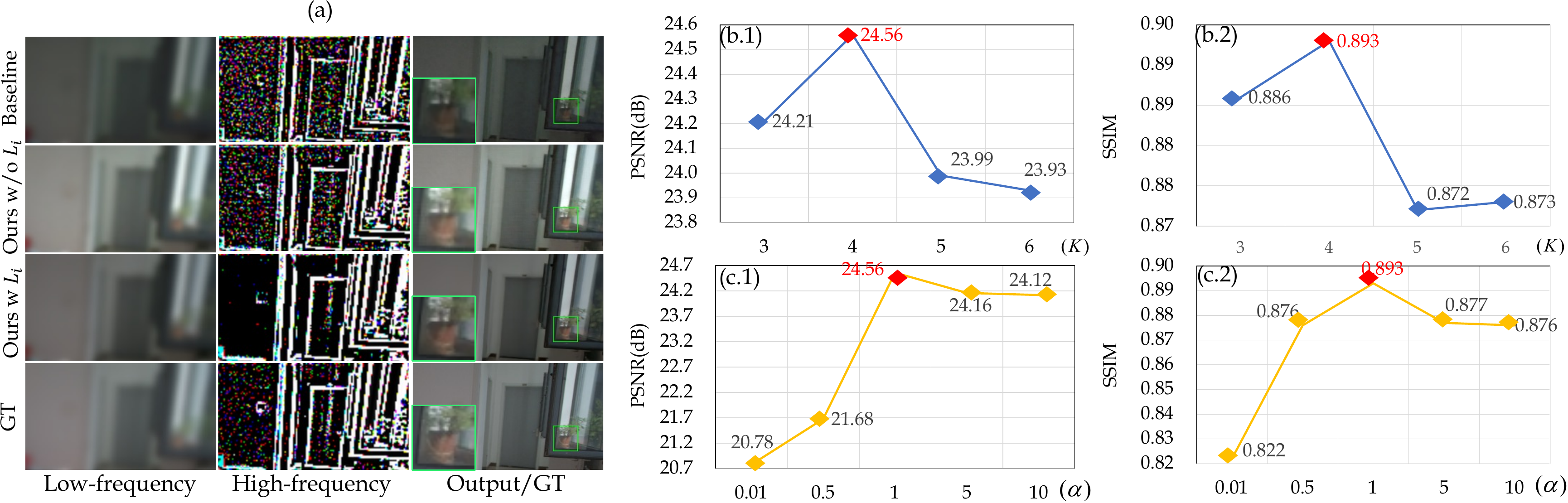} 
		\caption{(a) Visual validation of our disentanglement learning. (b-c) Impact of adopting different settings in our method: (b.1-2) different pyramid levels ($K$) in Laplace representation; (c.1-2) different $\alpha$ values to conduct low-frequency consistent supervision.} 
		
		\label{fig:ablation_exp}
		
	\end{figure} %

	\subsection{Ablation Studies}
	
	To verify the effectiveness of our proposed disentanglement learning scheme and ACCA, we conduct a comprehensive ablation analysis on the LOL-v2 dataset. Note that Restormer~\cite{zamir2022restormer} is chosen as our baseline model.
	
	\begin{table}[t]
		\setlength{\tabcolsep}{3pt}
		\begin{center}
			\caption{Effectiveness of our ACCA and $L_i$.}
			\label{table:decom}
			\begin{tabular}{l|c|c|c|c|c } 
				\hline
				
				Model &Restormer & ACCA &$L_i$  &PSNR/SSIM &Param./FLOPS\\
				\hline
				Baseline &\cmark  & &  &19.94/0.827 &26.13M/144.25G \\ 
				\rowcolor{gray!10}	Ours w/o $L_i$ &\cmark  &\cmark &  &20.21/0.837 &26.22M/146.78G \\ 
				\rowcolor{gray!10}	Ours w/ $L_i$ &\cmark  &\cmark &\cmark  &\textbf{24.56/0.893} &26.22M/146.78G \\ \hline
			\end{tabular}
		\end{center}
		
	\end{table}

	{\noindent\bf{Impacts of our designs.}}
	Compared to existing unified frameworks in SOTA LLIE methods, we propose a universal frequency-based disentanglement model and an efficient ACCA module for consistent coarse-to-fine enhancement. To evaluate our designs, we follow a systematic procedure, as shown in Tab.~\ref{table:decom}. First, we train a baseline model with the original unified framework. Next, we introduce the ACCA and LDRM modules, ablating the low-frequency consistent constraint between them. Finally, we train our full model with the proposed low-frequency consistency constraint.
	The results in Tab.~\ref{table:decom} and Fig.~\ref{fig:ablation_exp}(a) show that both the ACCA module and the disentanglement algorithm enhance performance. However, our full framework without the low-frequency consistent loss $L_i$ only improves by 0.27dB over the baseline, highlighting the challenges of unified enhancement. Additionally, our designs require minimal computational demand while significantly outperforming the baseline method.
	

	{\noindent\bf{Pyramid levels~($K$) in eq.~\ref{eq:lapde}.}}
	In LDRM, we introduce an image Laplace representation method to decompose images into high- and low-frequency components, using four pyramid levels by default. To explore the impact of different pyramid levels, we train additional models with 3, 5, and 6 levels. The results in Fig.~\ref{fig:ablation_exp} (b.1-2) show that our default setting provides the best performance.

	{\noindent\bf{Why not end-to-end?}}
	As previously discussed, prior approaches may encounter challenges in optimizing low-frequency and high-frequency components simultaneously, possibly resulting in suboptimal outcomes. To validate our claim, we conduct experiments using six baselines, training their enhanced versions with two different optimization strategies: (1) an end-to-end training scheme (where the ACCA's weights are not frozen) and (2) freezing the weights of the ACCA (our default decoupled optimization). From Tab.~\ref{table:freezeacca}, we can see that our decoupled optimization scheme consistently performs better than end-to-end learning.
	
	\begin{table}[h]
		\footnotesize
		\centering
		\caption{Quantitative comparison between two different optimization schemes. }
		\label{table:freezeacca}
		\setlength{\tabcolsep}{0.5pt}
		\begin{tabular}{l|c|c|c|c|c|c } 
			\hline
			Baseline &Retinexformer &Restormer &MIRNet &SNR  &LLFlow &Diff-L  \\ \hline
			End-to-end  &24.13/0.878 &24.27/0.885 &24.05/0.880 &23.82/0.864  &27.94/0.896 &23.90/0.847 \\ 
			\rowcolor{gray!10}
			Ours &24.21/0.881 &24.56/0.893 &24.19/0.882 &24.00/0.872  &28.90/0.908 &23.93/0.853 \\ \hline    
		\end{tabular}
	\end{table}
	
	{\noindent\bf{Different $\alpha$ values in Eq.~\ref{eq:loss}.}}
	For decoupled frequency optimization, we introduce a novel low-frequency consistent supervision. By default, we set the value of $\alpha$ to 1.0, which balances both the reconstruction and low-frequency consistent loss. In this experiment, we further explore the impact of different $\alpha$ values on the training of our LDRM. The quantitative results shown in Fig.~\ref{fig:ablation_exp}(c.1-2) demonstrate that an appropriate value of $\alpha$ leads to improved performance.

	\subsection{Limitation and future work.} While our frequency disentanglement optimization demonstrates significant enhancement over current state-of-the-art LLIE methods, it is confined to generating high-quality details that may be severely corrupted in the inputs. In the future, we intend to explore the incorporation of generative methods (such as Stable Diffusion models) into our framework.

	\section{Conclusion}
	In this study, we propose a novel framework for disentanglement learning in low-light image enhancement. Our approach involves explicitly decoupling the challenging LLIE problem into coarse low-frequency adjustment and fine high-frequency restoration tasks. To accomplish this, we introduce ACCA, a lightweight coarse adjustment module, that utilizes an efficient convolution composition module to achieve adaptive spatial-channel aggregation. Additionally, we develop LDRM by incorporating minimal re-designed SOTA LLIE methods. To ensure effective frequency decomposition, we introduce a low-frequency consistent loss for decoupled high-frequency restoration. Through comprehensive experiments, we demonstrate that our method significantly improves existing LLIE methods with minimal additional computational costs~(88K learnable parameters, 2.53GFLOPS for $256\times256$ images). 
	\\
	\\
	{\noindent\bf{Acknowledgments.}} This work is partially supported by Shenzhen Science and Technology Program KQTD20210811090149095 and also the Pearl River Talent Recruitment Program 2019QN01X226. The work was supported in part by the Basic Research Project No. HZQB-KCZYZ-2021067 of Hetao Shenzhen-HK S\&T Cooperation Zone, Guangdong Provincial Outstanding Youth Fund (No. 2023B1515020055), the National Key R\&D Program of China with grant No. 2018YFB1800800, by Shenzhen Outstanding Talents Training Fund 202002, by Guangdong Research Projects No. 2017ZT07X152 and No. 2019CX01X104, by Key Area R\&D Program of Guangdong Province (Grant No. 2018B030338001) by the Guangdong Provincial Key Laboratory of Future Networks of Intelligence (Grant No. 2022B1212010001), and by Shenzhen Key Laboratory of Big Data and Artificial Intelligence (Grant No. ZDSYS201707251409055). It is also partly supported by NSFC-61931024, NSFC-62172348, and Shenzhen Science and Technology Program No. JCYJ20220530143604010.
	
	\clearpage  

	%
	%
	\bibliographystyle{splncs04}

\end{document}